\documentclass[twocolumn,english]{IEEEtran}
\usepackage[T1]{fontenc}
\usepackage[latin9]{inputenc}
\usepackage{array}
\usepackage{multirow}
\usepackage{amsmath}
\usepackage{graphicx}

\makeatletter

\providecommand{\tabularnewline}{\\}

\@ifundefined{showcaptionsetup}{}{%
 \PassOptionsToPackage{caption=false}{subfig}}
\usepackage{subfig}
\makeatother

\usepackage{babel}
\begin{document}
\author{ 
\begin{tabular}[t]{c@{\extracolsep{8em}}c}Akash Chandrashekar, John Papadakis, Andrew Willis & Jamie Gantert \\ Department of Electrical and Computer Engineering & Munitions Directorate \\ University of North Carolina at Charlotte & Air Force Research Laboratory \\ Charlotte, NC 28223 & Eglin AFB, Florida 32542 \\ Email: achandr9@uncc.edu & Email: jamie.gantert.1@us.af.mil \end{tabular} }
\IEEEoverridecommandlockouts 
\makeatletter 
\let\ORGps@IEEEtitlepagestyle\ps@IEEEtitlepagestyle \def\ps@IEEEtitlepagestyle{\ORGps@IEEEtitlepagestyle \def\@oddfoot{\hbox{}\@IEEEfooterstyle\footnotesize \raisebox{\footskip}[0pt][0pt]{\@IEEEpubid}\hss\hbox{}}\relax 
\def\@evenfoot{\@oddfoot}} 
\makeatother 
%\IEEEpubid{978-1-5386-6133-8/18/\$31.00 ~\copyright2018 IEEE} 
\IEEEpubidadjcol

\title{Structure-From-Motion and RGBD Depth Fusion}
\maketitle
\begin{abstract}
This article describes a technique to augment a typical RGBD sensor
by integrating depth estimates obtained via Structure-from-Motion
(SfM) with sensor depth measurements. Limitations in the RGBD depth
sensing technology prevent capturing depth measurements in four important
contexts: (1) distant surfaces (>5m), (2) dark surfaces, (3) brightly
lit indoor scenes and (4) sunlit outdoor scenes. SfM technology computes
depth via multi-view reconstruction from the RGB image sequence alone.
As such, SfM depth estimates do not suffer the same limitations and
may be computed in all four of the previously listed circumstances.
This work describes a novel fusion of RGBD depth data and SfM-estimated
depths to generate an improved depth stream that may be processed
by one of many important downstream applications such as robotic localization
and mapping, as well as object recognition and tracking. 
\end{abstract}

\section{Introduction}

RGBD sensors are a relatively new class of image sensors. Their key
novel feature is the ability to simultaneously capture color ``RGB''
images of the scene and depth ``D'' images of scene- hence the term
``RGBD.'' RGB images are captured using a conventional visible light
camera that incorporates a lens to focus light rays from scene locations
onto distinct light-sensitive pixels of image sensor. Structured light
RGBD sensors consist of three integrated devices: an infrared (IR)
projector, an IR camera and an RGB camera in a rigid relative geometry
to create a single sensor that captures color-attributed $(X,Y,Z)$
surface data at ranges up to \textasciitilde{}6m with frame rates
up to 30 Hz. RGBD sensors have a wide range of applications which
include mapping, localization, pose estimation, and object recognition.
They have become popular for their ease-of-use and low cost in comparison
with other visual sensor technologies such as LIDAR, and have been
incorporated into consumer products like mobile phones, gaming consoles,
and automobiles \cite{litomisky2012consumer}.

\subsection{RGBD Sensing Technology and Limitations}

Depth image formation is accomplished using structured light technology
to measure the geometric position of viewed surfaces. This is accomplished
by illuminating scene surfaces with an infrared (IR) projector having
a known pattern and then using an IR camera to capture the projected
pattern \cite{Zhang:2012:MKS:2225053.2225203}. Deformation of the
projected pattern over scene object is analyzed and used to triangulate
the depth of scene surfaces with respect to the camera's optical axis.
The IR projector operates outside of visible light frequencies and,
as such, does not interfere with the captured RGB stream pixel values. 

Despite the popularity of RGBD sensors, their utility for generic
depth measurement is limited in several ways due to shortcomings associated
with structured light depth estimation. One significant shortcoming
is that RGBD sensors often fail to provide meaningful depth values
in sunlit outdoor scenes. Here the infrared radiation from sunlight
interferes with the projected pattern, causing the depth estimation
process to fail. This phenomenon also occurs in sunlit indoor scenes.
RGBD sensors also fail to collect measurements from surfaces having
specific reflectance properties. This includes the following three
reflectance contexts: (1) ``dark'' surfaces, i.e., surfaces having
a low reflectance, (2) specular, i.e., mirror-like, surfaces and (3)
transparent surfaces\textbf{ }\cite{8211432}\cite{Kadambi2014}.
All of these cases represent instances where the projected IR pattern
reflected into the IR camera is not observable from the background
due to signal interference, e.g., sunlight, or weak/dim images, e.g.,
low reflectance, mirror-like and transparent surfaces.

\subsection{Structure from Motion }

The Structure from Motion (SfM) algorithm leverages ideas originally
drawn from photogrammetry to estimate the three-dimensional structure
of a scene from a time series of RGB images from a moving single camera.
This is achieved by calibrating the camera \cite{Zhang:2000:FNT:357014.357025}
to develop a highly-accurate model to describe how 3D positions are
projected into camera images. Using this image formation model, the
SfM algorithm matches together pixels in separate images that correspond
to projections of the same 3D location as the camera moves in the
scene. Using the camera projection model and the assumption that matched
pixels are measurements of the same 3D world coordinates, the SfM
algorithm solves for both the pose of the camera within the global
coordinate system and the set of 3D surface positions provided by
corresponding image pixels \cite{6751290}. The SfM problem is non-linear
in the unknowns and is typically solved in a two-stage sequence. Stage
1 solves for the relative pose of the camera at the instant the images
were recorded. Stage 2 conditions on the estimated camera pose values
and solves for the 3D scene structure. Both stages use correspondences
between pixels from different images to solve the non-linear equations
in the unknown variables. The camera pose tracking problem of Stage
1 is often solved by finding a map that transforms pixels from the
original $(x,y)$ coordinate field to new coordinate positions $(x',y')$
such that both locations correspond to images of the same 3D scene
point. The multi-view 3D surface reconstruction of Stage 2 is often
solved using the bundle adjustment algorithm \cite{Triggs:1999:BAM:646271.685629}.
\begin{figure*}
\noindent \centering{}\subfloat[]{\noindent \begin{centering}
\includegraphics[height=1.3in]{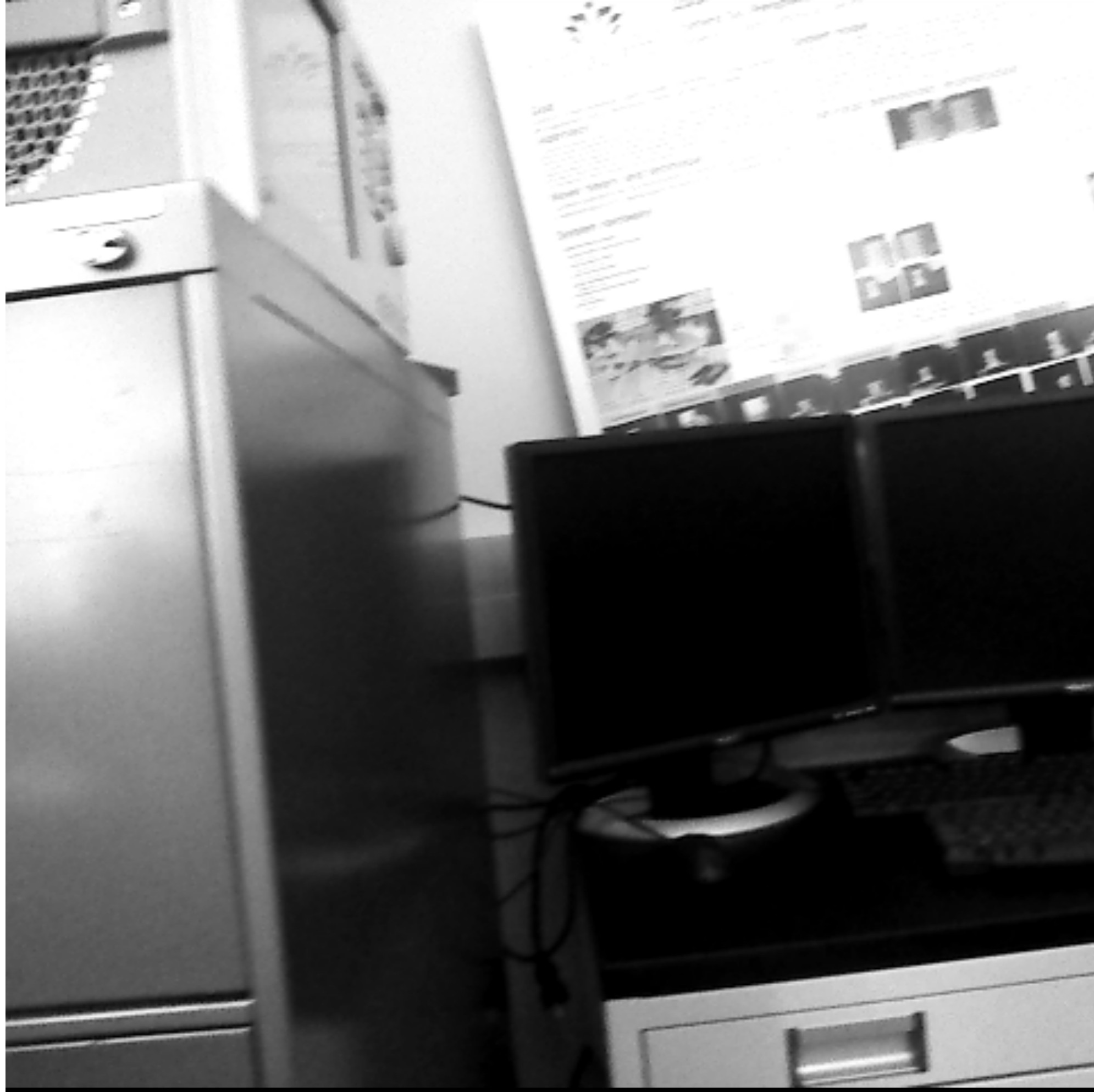}
\par\end{centering}
}\subfloat[]{\noindent \begin{centering}
\includegraphics[height=1.3in]{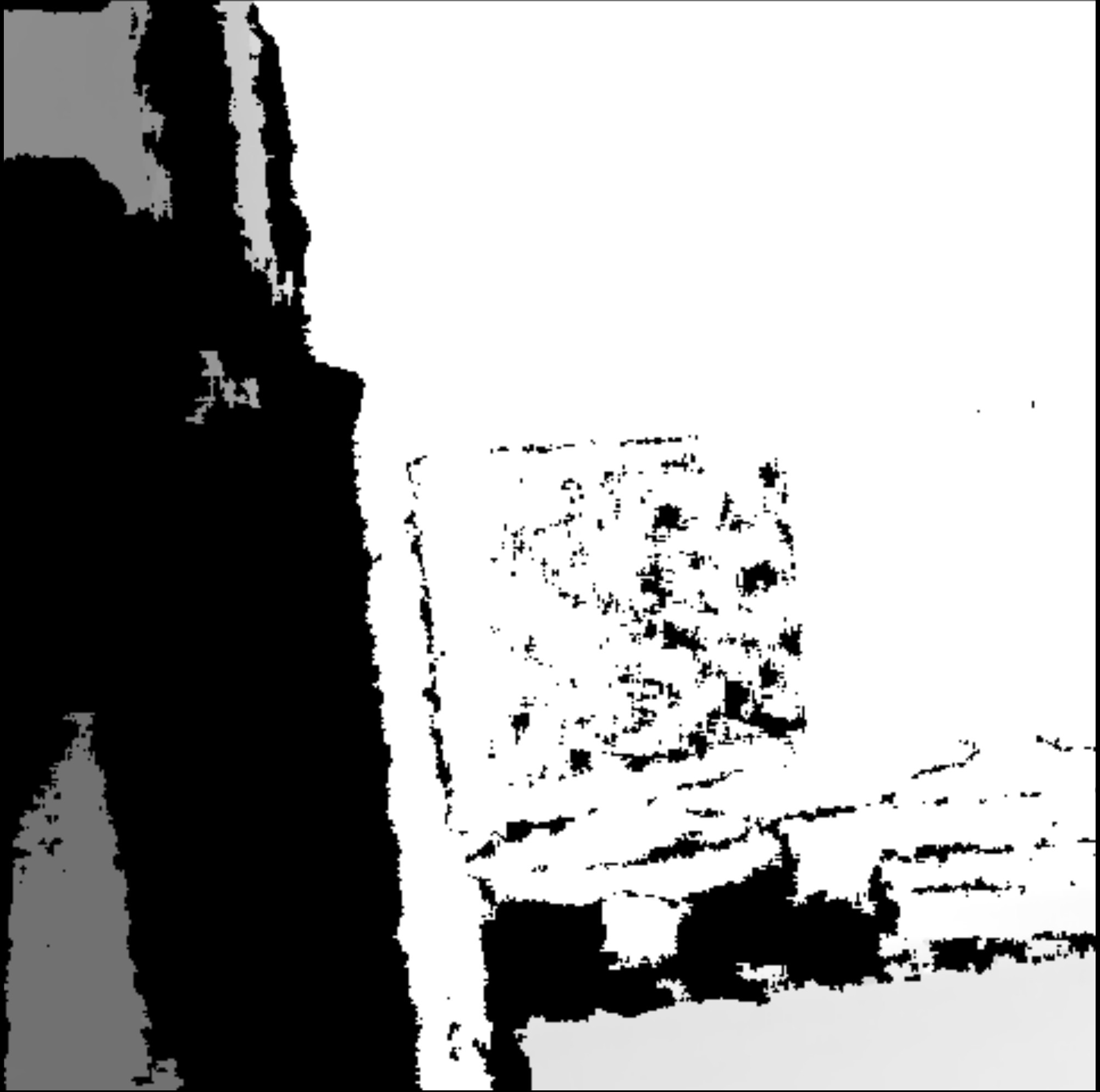}
\par\end{centering}
}\subfloat[]{\noindent \begin{centering}
\includegraphics[height=1.3in]{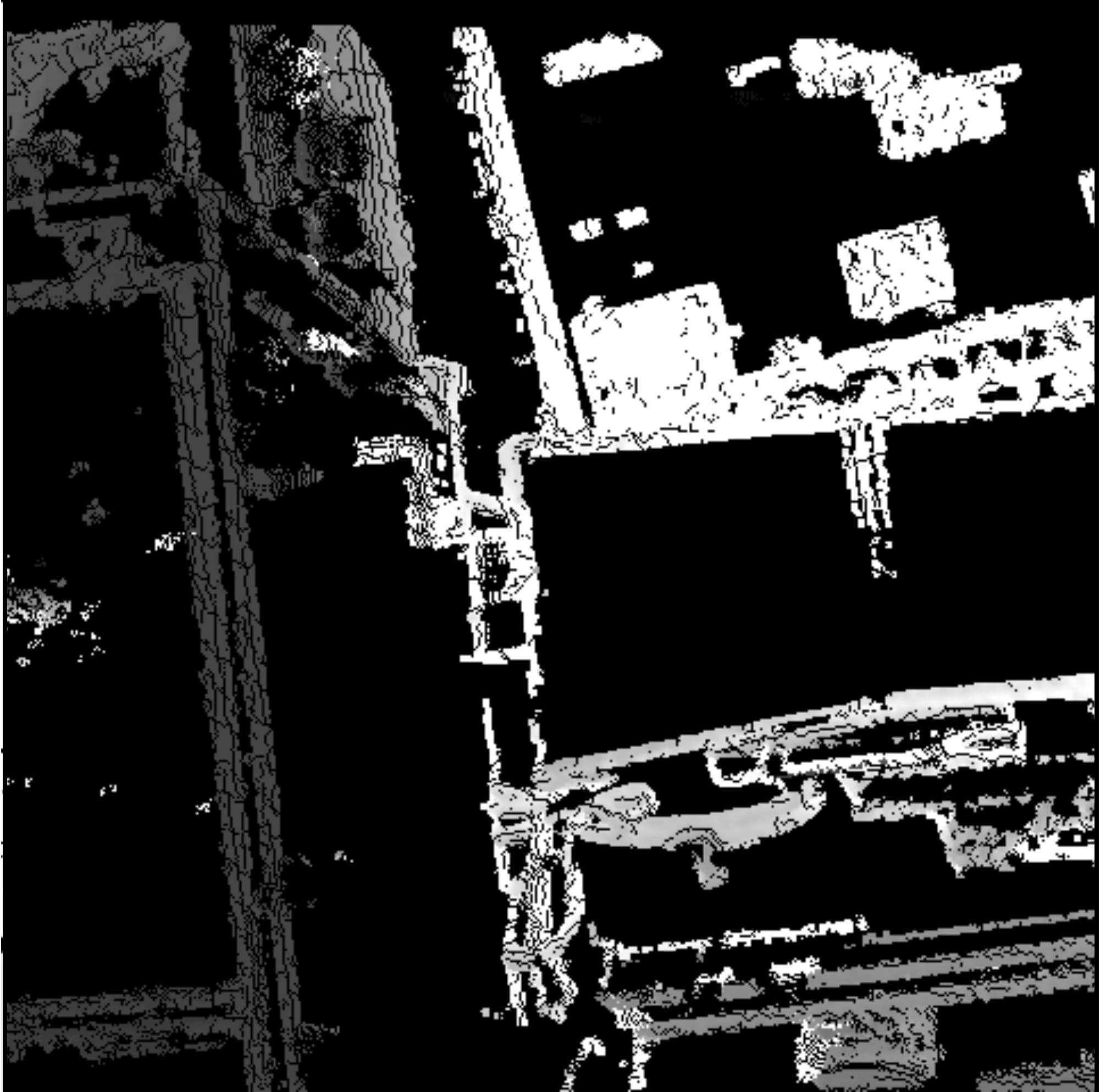}
\par\end{centering}
}\subfloat[]{\noindent \begin{centering}
\includegraphics[height=1.3in]{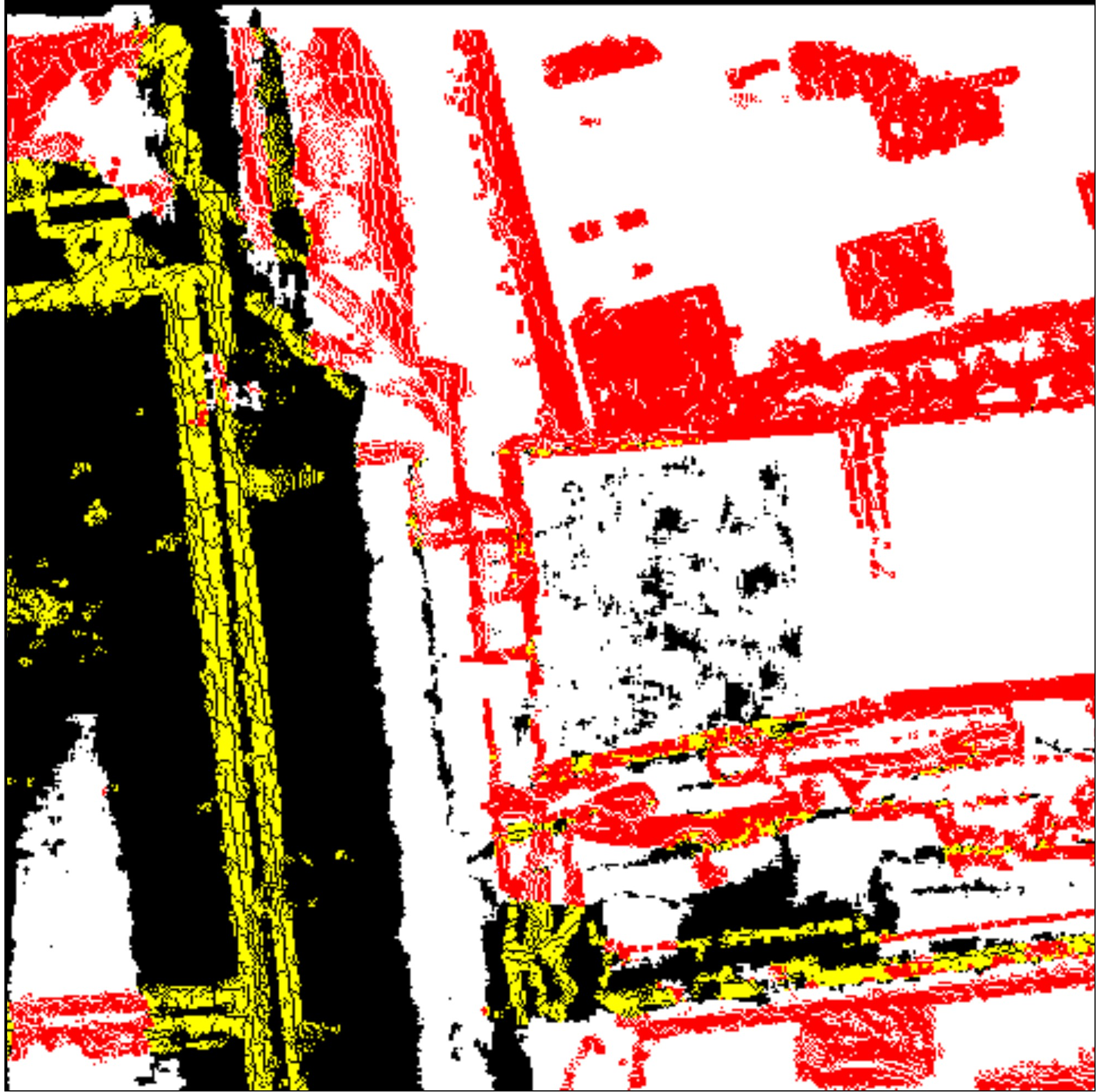}
\par\end{centering}
}\caption{\label{fig:gussian fussion}An overview of the proposed RGBD and SfM
fusion algorithm (a) shows a grayscale image of the scene (b) shows
the sensed RGBD depth image (c) shows the SfM-estimated depth image
and (d) shows the fused image. The fused image has been color-coded
as follows: (white) denotes depth locations sensed only by the RGBD
sensor, (yellow) denotes depth locations only sensed via SfM, (red)
denotes fused (RGBD+SfM) depth locations and (black) denotes depth
locations without RGBD or SfM measurements.}
\end{figure*}

While scene reconstruction via SfM produces depth images in contexts
where depth cameras fail, this modality for depth estimation also
has several shortcomings. Specifically, the theoretical formulation
of the SfM problem shows that the scale of the estimated 3D structure
cannot be known without prior or outside information. This complicates
both the mathematical and computational SfM solutions. SfM also presumes
that viewed surfaces are static, i.e., they do not move, and when
this assumption is violated reconstructed surfaces are highly inaccurate.

\subsection{Contribution}

This article seeks to leverage the strengths of structured light sensor
derived and SfM derived depth measurements by fusing these measurements
into an improved depth image that provides depth measurements in contexts
where at least one of the two depth estimation approaches succeeds.
Figure \ref{fig:gussian fussion} shows an RGBD-SfM fusion result
for an indoor scene and how the fusion result (Figure \ref{fig:gussian fussion}(d))
captures more scene geometry than either approach independently. Our
proposed method to fuse RGBD and SfM depth imagery includes consideration
of the RGBD sensor depth noise model, the SfM algorithm depth noise
model and also copes with the inherent unknown scale and scale-drift
problems intrinsic to SfM. To our knowledge these technical issues
have not been discussed elsewhere in the literature.

\section{Related Work and Background Information}

This article proposes fusion of SfM-estimated depths with depths captured
from an RGBD image sensor. This section is dedicated to discussing
the relevant aspects of the SfM algorithm and the sensed RGBD measurements
necessary to explain the proposed fusion method. Specifically, this
section reviews the theoretical details of the SfM algorithm, methods
for processing depth images including computing depth images for arbitrary
camera poses, and details existing knowledge regarding the sensor
measurement noise for RGBD depth measurements.

\subsection{SfM Depth Reconstruction}

The SfM algorithm uses a time sequence of images from a moving camera
to recover the 3D geometry of objects viewed by the camera. While
this problem can be solved without a calibrated camera, reconstruction
accuracy will adversely affected. This work assumes that the camera
calibration \cite{5534797} parameters are known. The SfM algorithm
can be broken down into two key steps:
\begin{enumerate}
\item Estimation of the camera pose, i.e., position and orientation, at
the time each image was recorded,
\item Estimation of the 3D structure of the scene.
\end{enumerate}
As previously mentioned, typical SfM systems solve (1) by computing
a map that associates pixels from the original $(x,y)$ coordinate
field to new coordinate positions $(x',y')$ such that both locations
correspond to images of the same 3D scene point and (2) via multi-view
3D surface reconstruction algorithm, e.g., bundle adjustment \cite{Triggs:1999:BAM:646271.685629}.
In the following sections we provide an overview of aspects of the
SfM algorithm necessary for the development of the proposed RGBD-SfM
depth fusion algorithm.

\subsubsection{Solving for Image Pixel Correspondences}

There are generically two different approaches for finding corresponding
observations of the same 3D surface location in multiple images referred
to as \emph{direct }and \emph{indirect }\cite{7898369}\emph{. }In
this discussion, we refer to the correspondence problem as a source-to-target
matching problem. Let $\mathbf{I}_{t}(x,y)$ denote an image recorded
at time $t$ and $\mathbf{I}_{t+\Delta t}(x,y)$ denote a subsequent
image measured at time $t+\Delta t$. The correspondence problem seeks
to find a map that transforms pixels from the original $(x,y)$ coordinate
field of $\mathbf{I}_{t}$ to new coordinate positions $(x',y')$
in $\mathbf{I}_{t+\Delta t}$ such that $\mathbf{I}_{t}(x,y)$ and
$\mathbf{I}_{t+\Delta t}(x',y')$ correspond to projections of the
same 3D scene point. 

\subsubsection*{Indirect Methods}

Indirect methods compute this mapping by detecting special $(x,y)$
locations referred to as feature locations with a purpose-built feature
detection algorithm, e.g., the Harris corner detector \cite{Harris88acombined}.
A description of the image patch in the vicinity of each detected
$(x,y)$ location is computed using some feature descriptor algorithm,
e.g., Lowe's SIFT descriptor \cite{Lowe2004}. Feature descriptors
seek to provide a vector of values from the image patch data that
is invariant to the image variations that occur during camera motion.
These include but are not limited to the following effects: illumination
variation, affine and/or projective invariance, photometric invariance
(brightness constancy), and scale invariance. 

Popular feature descriptors often prioritize scale and affine invariance
as their strengths. The invariance property allows for correspondences
to be computed by finding the mapping from the feature descriptor
set calculated from image $\mathbf{I}_{t}(x,y)$ to the feature descriptor
set calculated from image $\mathbf{I}_{t+\Delta t}(x,y)$. Solutions
often assume the map solution is 1-to-1, i.e., a single 3D location
can only project to one location in each image, and the descriptor
values are invariant across the image pair. As such, solving for the
correspondence reduces to an assignment problem where we seek to find
the correspondence that minimizes the difference between corresponding
descriptors. Let $\mathbf{d}_{i,t}$ denote the $i^{th}$ descriptor
from $\mathbf{I}_{t}(x,y)$ and $\mathbf{d}_{j,t+1}$ denote the $j^{th}$
descriptor from $\mathbf{I}_{t+\Delta t}(x,y)$. We then specify the
indirect correspondence problem as a search for the correspondence
set ${\cal C}$ consisting of index pairs $(i,j)$ that minimize the
total descriptor error as shown in equation (\ref{eq:correspondence})
with the standard squared vector norm error function shown in equation
(\ref{eq:vector_sqnorm_errorfunc}).

\begin{equation}
\widehat{{\cal C}}=\min\sum_{(i,j)\in{\cal C}}e(\mathbf{d}_{i,t},\mathbf{d}_{j,t+1}),\label{eq:correspondence}
\end{equation}
\begin{equation}
e(\mathbf{d}_{i,t},\mathbf{d}_{j,t+1})=\left\Vert \mathbf{d}_{i,t}-\mathbf{d}_{j,t+1}\right\Vert ^{2}\label{eq:vector_sqnorm_errorfunc}
\end{equation}

\subsubsection*{Direct Methods}

Direct methods on the other hand typically iteratively solve for a
set of transformation parameters that best align a pair of images
by the minimization of pixel-wise errors. An image warping function,
$\omega(\mathbf{x},\xi)$, maps a pixel location, $\mathbf{x}=[x,y]^{t}$,
in the original coordinate field to new coordinate positions, $\mathbf{x}'$,
such that both locations correspond to images of the same 3D scene
point. A classical solution to this problem is given by the Lucas-Kanade-Tomassi
(LKT) camera tracking algorithm \cite{Baker:2004:LYU:964568.964604}. 

This is analogous to the process of image alignment, whose goal is
to minimize the sum of squared error between two images, the template
image $\mathbf{T}(\mathbf{x})$, and the moving image, $\mathbf{I}(\mathbf{x})$.
In this method the moving image is mapped onto the coordinate frame
of the template image using a warp function, $\omega(\mathbf{x},\xi)$,
as in equation \ref{image-alignment}.

\begin{equation}
\widehat{\xi}=\min_{\xi}\underset{\mathbf{x}}{\sum}\left(\mathbf{I}(\omega(\mathbf{x},\xi))-\mathbf{T}(\mathbf{x})\right)^{2}\label{image-alignment}
\end{equation}

In equation (\ref{image-alignment}), the warp function, $\omega(\mathbf{x},\xi)$,
maps pixel locations, $\mathbf{x}$, in the template to pixel locations
in the image $\mathbf{I}(\mathbf{x})$ using the warp transformation
parameters $\xi$. For direct correspondence estimation, $\xi$ is
a pose transformation of the viewing camera represented as a unknown
6x1 vector. We then seek the camera pose transformation parameters,
$\widehat{\xi}$, that minimize equation (\ref{image-alignment})
which provides the camera pose change that best explain the differences
in these two of images of the same scene. Given two images and the
camera pose change between them, one can take the information in one
image, and through the warp function map these values into the viewpoint
of the other image to establish a correspondence. In this case the
theoretical difference between expected and observed values for the
image pair is zero if the camera pose change is known exactly and
sensor noise and other outside influences are ignored. 

\subsubsection{Solving for Scene Geometry}

There are generally two different approaches for 3D reconstruction
referred to as \emph{sparse} and \emph{dense} approaches. Sparse methods
reconstruct the 3D scene geometry only for a select subset of the
entire image data \cite{klein07parallel}. This subset is often corner
locations or locations marked by some type of extracted feature, e.g.,
SIFT or SURF. \cite{Lowe2004,Bay:2008:SRF:1370312.1370556} This results
in a sparse description of the 3D scene in terms of a sparse cloud
of 3D points. In contrast, dense methods \cite{6696650} reconstruct
as many 3D geometric locations as possible and seek to provide a more
complete description of the scene.

Sparse reconstructions often benefit from having a lower computational
cost but provide few 3D measurements. Dense reconstructions have higher
computational cost but provide a much more complete description of
the 3D scene. Dense reconstruction techniques have seen much recent
interest, although a highly accurate, dense, and real-time SfM approach
has remained elusive. 

A third class of algorithms, referred to as \emph{semi-dense} algorithms
\cite{6751290}, seek to strike a compromise between the sparse and
dense methods. The reconstruction techniques used are most similar
to dense methods, however, only a subset of all image pixels are reconstructed.
These approaches leverage the high accuracy of dense reconstruction
techniques, but are sparse enough to allow for real-time operation.

Here, reconstruction is limited to those pixels which possess high
intensity gradient values. These regions often correspond to scene
geometries such as edges, corners, and curves and to other areas of
the scene that are highly textured. The thought here is that regions
of the image that possess large changes in intensity convey more information
than regions that possess less, thus semi-dense reconstructions provide
a compressed version of the total dense scene structure.
\begin{figure*}
\noindent \begin{centering}
\includegraphics[width=15cm]{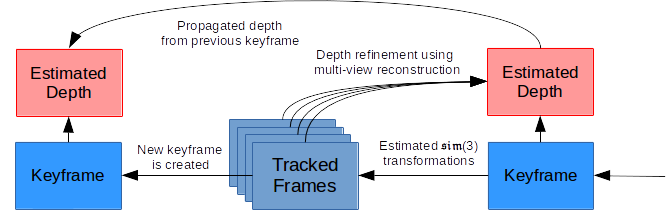}
\par\end{centering}
\caption{\label{fig:LSDSLAM-Overview} An overview of the LSD SLAM algorithm.
Each incoming frame is tracked against the current keyframe. If it
does not satisfy the critera for new keyframe creation, it is used
along with previous tracked frames for refinement of the estimated
depth values of the keyframe. Otherwise, the frame is considered a
new keyframe and depth estimates from the previous keyframe are propagated
and used for initialization of the new depth estimates.}
\end{figure*}

\subsubsection{LSD SLAM}

We leverage the SfM algorithm proposed by Engel et al. dubbed LSD-SLAM.
\cite{Engel2014} This approach is a \emph{semi-dense}, \emph{direct}
method which optimizes the geometry directly on the image intensities.
LSD-SLAM provides as output a reconstruction of the observed 3D environment
as a pose-graph of specially designated RGB images called ``keyframes''
with associated semi-dense depth images. Direct correspondences are
found between every RGB frame and each RGB keyframe to estimate the
keyframe-to-RGB-frame relative camera pose. This is achieved by Levenberg-Marquardt
optimization of the photometric reprojection error between the RGB
image pair.

\begin{equation}
E(\xi)=\underset{i}{\sum}(I_{ref}(P_{i})-I(\omega(p_{i},D_{ref}(p_{i}),\xi)))^{2}\label{eq:Photomatric Error}
\end{equation}

Equation (\ref{eq:Photomatric Error}) shows the specific image alignment
objective function and formalizes the form for equation (\ref{image-alignment}).
Here, the warp function relies on the current estimate of depth, $D_{ref}$,
to determine the relative pose change as a similarity transform: $\xi\in sim(3)$.
The reference depth, $D_{ref}$, is the depth associated with current
keyframe, these depth values could be initialized either with random
values or with the depth measurements from a RGBD sensor to initiate
the process. For every new frame tracked, the depths associated with
the keyframe are continuously refined by performing adaptive baseline
stereo 3D reconstruction \cite{6751290} using the newly tracked frame
along with a stack of frames that have been tracked previously. When
there is drastic pose change between the tracked frame and keyframe,
the current tracked frame is promoted to a keyframe and the depth
map from the previous keyframe is propagated to the new keyframe using
the estimated pose change and regularized. All keyframes are added
as nodes to a pose graph that stores the relative pose between the
keyframes as edges/constraints \cite{5979949}. The pose graph stores
the global trajectory of the camera in the 3D scene and a graph optimization
algorithm processes the pose graph to improve the camera pose estimates
as new image correspondences are found by image matching and loop
closures. 

\subsection{Depth Image Processing}

Depth fusion requires a significant amount of depth image processing.
This section details how depth images can be used to reconstruct 3D
point clouds and how point cloud re-projection can be used to predict
the depth image captured from a camera in an arbitrary target pose
from a measured depth image and knowledge of the relative pose between
the the measured image and the target pose.

\subsubsection{Point Cloud Reconstruction\label{subsec:Point-Cloud-Reconstruction}}

Measured 3D $(X,Y,Z)$ positions of sensed surfaces can be directly
computed from the intrinsic camera parameters and depth image values.
Here, the $Z$ coordinate is directly taken as the depth value and
the 3D $(X,Y)$ coordinates are computed using the pinhole camera
model. In a typical pinhole camera model 3D $(X,Y,Z)$ points are
projected to $(x,y)$ image locations \cite{Ma:2003:IVI:971144},
e.g., for the image columns the $x$ image coordinate is $x=f_{x}\frac{X}{Z}+c_{x}-\delta_{x}$.
However, for a depth image, this equation is re-organized to ``back-project''
the depth into the 3D scene and recover the 3D $(X,Y)$ coordinates
as shown by equation (\ref{eq:RGBD_reconstruction}) 

\begin{equation}
\begin{array}{ccc}
X & = & (x+\delta_{x}-c_{x})Z/f_{x}\\
Y & = & (y+\delta_{y}-c_{y})Z/f_{y}\\
Z & = & Z
\end{array}\label{eq:RGBD_reconstruction}
\end{equation}
where $Z$ denotes the sensed depth at image position $(x,y)$, $(f_{x},f_{y})$
denotes the camera focal length (in pixels), $(c_{x},c_{x})$ denotes
the pixel coordinate of the image center, i.e., the principal point,
and $(\delta_{x},\delta_{y})$ denote adjustments of the projected
pixel coordinate to correct for camera lens distortion.

\subsubsection{Point Cloud Re-Projection\label{subsec:Point-Cloud-Re-Projection}}

Depth images can be simulated for camera sensor in arbitrary poses
by applying back-projection, transformation on the resulting points,
and projection into the new coordinate frame. For discussion, assume
that depth image $\mathbf{d}(x,y)$ has been recorded in the ``standard''
camera/optical coordinate system where the origin corresponds to the
camera focal point, the $z$-axis corresponds to the depth/optical
axis extending out into the viewed scene, the $x$-axis points towards
the right and spans the image columns and the $y$-axis points downward
and spans the image rows. 

Let $\mathbf{R}$ denote the 3D rotation that rotates the coordinate
axes of the standard coordinate system to align with the same axes
of a second camera having arbitrary pose. Similarly, Let $\mathbf{t}$
denote the 3D translation vector describing the position of the focal
point of a second camera having arbitrary pose. Using this notation,
the depth transformation can be accomplished in the following three
steps:
\begin{enumerate}
\item Back-project $\mathbf{d}(x,y)$ to create an $(X,Y,Z)$ point cloud
(as described in $\S$~\ref{subsec:Point-Cloud-Reconstruction}), 
\item Transform, i.e., rotate and translate, each point $\mathbf{p}_{i}=[X,Y,Z]^{t}$
in the point cloud to generate a new point $\mathbf{p}_{i}^{'}=[X^{'},Y^{'},Z^{'}]^{t}$
that lies in a standard optical coordinate system centered on the
second camera's focal point and having orientation that aligns with
corresponding $x,y,z$-axes using equation (\ref{eq:coordinate_system_change}),
\begin{equation}
\mathbf{p}_{i}^{'}=\mathbf{R}(\mathbf{p}_{i}-\mathbf{t})\label{eq:coordinate_system_change}
\end{equation}
\item Project the $(X,Y,Z)$ point cloud using the pinhole camera equations
to compute the new depth image $\mathbf{d}^{'}(x,y)=Z$ using equation
(\ref{eq:re_projection}).
\end{enumerate}
\begin{equation}
\begin{array}{ccc}
x & = & f_{x}(\frac{X^{'}}{Z^{'}})-\delta_{x}+c_{x}\\
y & = & f_{y}(\frac{Y^{'}}{Z^{'}})-\delta_{y}+c_{y}\\
Z & = & Z^{'}
\end{array}\label{eq:re_projection}
\end{equation}

Typically, the projected point cloud measurements fall at non-integer
locations in the new depth image and the values of $\mathbf{d}^{'}(x,y)$
must then be interpolated using some interpolation scheme such as
nearest neighbor or bilinear interpolation. 

\subsection{RGBD Measurement Noise\label{subsec:RGBD_Measurement-Noise}}

The proposed fusion algorithm relies on experimental studies of accuracy
and noise for RGBD sensor measurement, e.g., the Kinect sensor. Research
in \cite{7925286} shows that a Gaussian noise model provides a good
fit to observed measurement errors on planar targets where the distribution
parameters are mean $0$ and standard deviation$\sigma_{Z}=\frac{m}{2f_{x}b}Z^{2}$
for depth measurements where $\frac{m}{f_{x}b}=-2.85e^{-3}$ is the
linearized slope for the normalized disparity empirically found in
\cite{7925286}. Since the 3D coordinates for $(X,Y)$ are a function
of both the pixel location and the depth, their distributions are
shown below:

\begin{equation}
\begin{array}{ccc}
\sigma_{X} & = & \frac{x_{im}-c_{x}+\delta_{x}}{f_{x}}\sigma_{Z}=\frac{x_{im}-c_{x}+\delta_{x}}{f_{x}}(1.425e^{-3})Z^{2}\\
\sigma_{Y} & = & \frac{y_{im}-c_{y}+\delta_{y}}{f_{y}}\sigma_{Z}=\frac{y_{im}-c_{y}+\delta_{y}}{f_{y}}(1.425e^{-3})Z^{2}\\
\sigma_{Z} & = & \frac{m}{f_{x}b}Z^{2}\sigma_{d'}=(1.425e^{-3})Z^{2}
\end{array}\label{eq:noise_models}
\end{equation}

These equations indicate that 3D coordinate measurement uncertainty
increases as a quadratic function of the depth for all three coordinate
values. However, the quadratic coefficient for the $(X,Y)$ coordinate
standard deviation is at most half that in the depth direction, i.e.,
$(\sigma_{X},\sigma_{Y})\approx0.5\sigma_{Z}$ at the image periphery
where $\frac{x-c_{x}}{f}\approx0.5$, and this value is significantly
smaller for pixels close to the optical axis. Variances derived from
this noise model can then be used to fuse depth measurements resulting
from structured light sensors with depth measurements from other sources.

\section{Methodology}

\begin{figure*}
\noindent \centering{}\includegraphics[width=7in,height=2.5in,keepaspectratio]{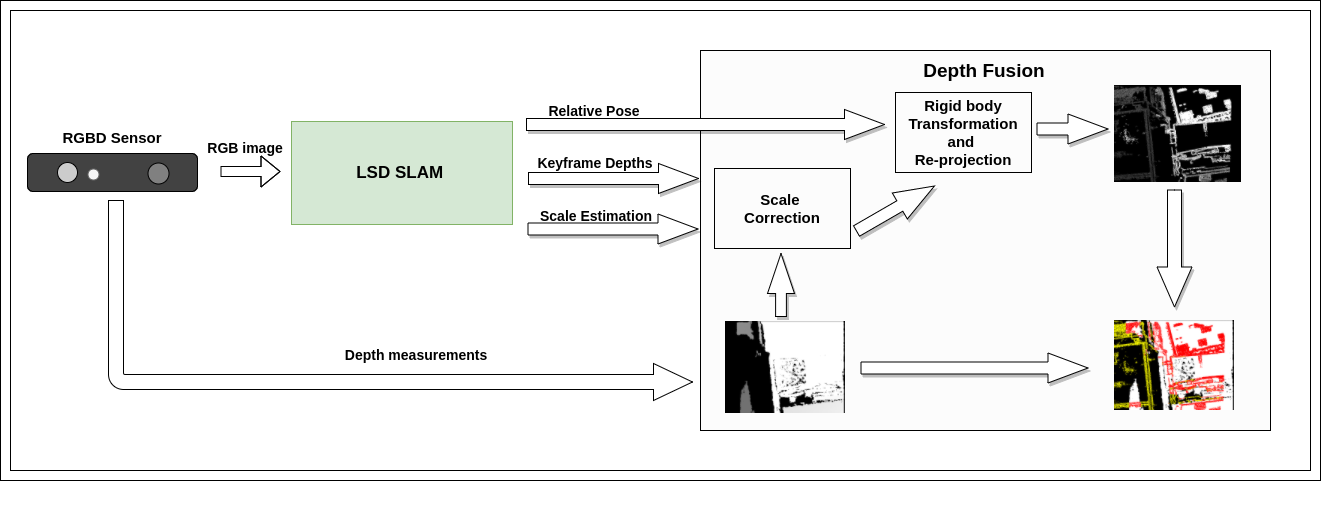}\caption{\label{fig:The-complete-process}An overview of the proposed depth
fusion algorithm.}
\end{figure*}
The proposed fusion approach applies the semi-dense monocular reconstruction
approach referred to as Large Scale Direct (LSD) SLAM \cite{Engel2014}.
The LSD-SLAM algorithm solves the SfM problem using a \emph{direct}
method to compute pixel correspondences and a \emph{semi-dense} method
for 3D reconstruction. We select this approach as it does not require
or impose any prior knowledge about the scene structure as required
by \emph{dense }reconstruction methods and it gives more 3D estimates
than \emph{sparse} approaches while having similar computational cost.

The LSD-SLAM SfM algorithm consists of the following three components:
\begin{enumerate}
\item A tracking component that estimates the pose of the camera
\item A depth map estimation component that estimates semi-dense depth images
for keyframes
\item A map optimization component that seeks to create a 3D map of the
environment that is self-consistent.
\end{enumerate}
This work utilizes the first two components to explore fusion of RGBD
depth images with SfM depth images. For this work, the map optimization
component (3) is not used. Figure \ref{fig:The-complete-process}
depicts an overview of the proposed depth fusion algorithm.

\subsection{Time and Spatial Sampling Issues\label{subsec:Time-and-Spatial}}

As mentioned previously, RGBD sensors measure depth at a rate of 30
frames per second (fps) and LSD-SLAM computes depth images only for
\emph{keyframes }which is a sparse subset of the measured RGB frames.
Further, keyframes are not generated uniformly in time but created
when the SfM algorithm detects criteria required to create a new keyframe.
This condition is triggered when the current camera pose is too far
from the most recent keyframe camera pose and when the current frame
tracking result is ``good'' in the sense that the image warping
correspondence objective function suggests an accurate or low-error
result. As a result, SfM-estimated depths exist only for those RGB
images designated as keyframes.

Further, the spatial distribution of SfM-estimated depths within SfM
keyframes are localized to only those pixels having ``good'' 3D
reconstruction characteristics. In this sense, the quality of the
depth estimate depends on accurately matching pixels along epipolar
lines inscribed in the image. The matching performance here is best
when there is a significant change in the image intensities along
the epipolar line. Hence, 3D depth reconstruction is limited to those
pixels that lie at sharp intensity changes, i.e., ``edge'' pixels,
and further limited to those ``edge'' pixels that lie on edges that
are roughly perpendicular to the direction of the epipolar line (see
\cite{6751290} for details). 

The LSD-SLAM algorithm estimates depth at ``good'' pixel positions
as a 1-dimensional Gaussian distribution specified as a mean image
$\mu_{SfM}(x,y)$, i.e., the estimated depth image, and a variance
image $\sigma_{SfM}^{2}(x,y)$ such that the RGB keyframe pixel at
location $\mathbf{I}(x,y)$ is estimated to have depth $\mu_{SfM}(x,y)$
with uncertainties given by $\sigma_{SfM}^{2}(x,y)$. In this sense,
the keyframe image $\mathbf{I}(x,y)$ augmented with the estimated
depth image $\mu_{SfM}(x,y)$ is analogous in format to sensed RGBD
image data. Yet, the uncertainties for the image $\mu_{SfM}(x,y)$
are given by the image $\sigma_{SfM}^{2}(x,y)$ rather than the experimentally
validated uncertainties discussed in \S~ \ref{subsec:RGBD_Measurement-Noise}.

\subsection{Image Registration Issues}

Fusing depth measurements requires knowledge of the correspondence
between the depth measurements generated from the RGBD sensor and
the SfM algorithm. For SfM keyframes this correspondence is trivial
due to the fact that RGBD sensors support hardware registration. Hardware
registration co-locates the RGBD depth image measurements, $\mathbf{d}_{rgbd}(x,y)$,
and RGB appearance values, $\mathbf{I}(x,y)$. Hence, for hardware-registered
RGBD depth images, $\mathbf{d}_{rgbd}(x,y)$ is the measured depth
of the surface having RGB pixel $\mathbf{I}(x,y)$. Similarly, SfM-estimated
depths for an RGB keyframe, $\mu_{SfM}(x,y)$, are the depths for
the surface having RGB pixel $\mathbf{I}(x,y)$. Hence fusion is accomplished
by fusing the measurements at corresponding $(x,y)$ locations in
the RGBD depth image, $\mathbf{d}_{rgbd}(x,y)$, and the SfM depth
image, $\mu_{SfM}(x,y)$.

Depth correspondences for RGB images that are not SfM keyframes must
be computed from one or more SfM keyframe depth images. This article
uses the most recent, i.e., closest-in-time, keyframe to generate
co-registered SfM depth images for arbitrary RGB images. To do so,
the depth image from the most recent keyframe, $\mu_{SfM}(x,y)$,
is back-projected \ref{subsec:Point-Cloud-Reconstruction} to create
a 3D point cloud of SfM measurements. Using the estimated keyframe-to-camera
pose change, the 3D measurements are transformed, then projected \ref{subsec:Point-Cloud-Re-Projection}
into the RGB camera image plane using the 3D projection equations
for the camera provided via camera calibration. The resulting depth
image, $\widetilde{\mu}_{SfM}(x,y)$ is then co-registered with the
RGBD depth image $\mathbf{d}_{rgbd}(x,y)$ and RGB image $\mathbf{I}(x,y)$.

Using these techniques, co-registered SfM depth images can be computed
for a general RGBD image. When RGBD images correspond to SfM keyframes
the registration is ``automatic,'' i.e., no computation is necessary.
In all other cases, a co-registered SfM depth image must be computed
by reconstructing a 3D point cloud from a keyframe and then projecting
the point cloud into the target RGB camera image using the estimated
camera calibration and keyframe-to-camera-frame relative pose parameters.

\subsection{Resolving the Unknown SfM Scale}

The methods described in previous sections detail how co-registered
SfM depth measurements are computed for every sensed RGBD frame. However,
as discussed in previously, SfM depth images intrinsically have an
unknown scale, $\alpha$, which reflects the fact that the solution
for the scene structure is not geometrically unique, i.e., the same
scene structure can be observed at a infinite number of distinct scales.
Therefore fusion requires the scale of the SfM depth image to fit
the scale of the real-world scene measured by the RGBD camera.

Given that the depth measurements for the RGBD depth image are co-registered
with the SfM estimated depth image the scale parameter can be directly
estimated by minimizing the sum of the squared depth errors \cite{Bishop:2006:PRM:1162264}
between the SfM depth image and the RGBD depth image. Let $V$ denote
the set of $(x,y)$ positions that have valid depth measurements for
``standard'' fusion as described in the section \ref{subsec:Fusing-RGBD-Depths}.
Equation (\ref{eq:scale error}) shows the error function used to
compute the unknown scale value and equation (\ref{eq:depth_scale_solution})
shows the solution $\widehat{\alpha}$ that minimizes this error.

\begin{equation}
e(\alpha)=\sum_{(x,y)\epsilon V}\left\Vert \mathbf{d}_{rgbd}(x,y)-\alpha\mu_{SfM}(x,y)\right\Vert ^{2}\label{eq:scale error}
\end{equation}

\begin{equation}
\widehat{\alpha}=\sum_{(x,y)\epsilon V}\frac{\mathbf{d}{}_{rgbd}(x,y)}{\mu_{SfM}(x,y)}\label{eq:depth_scale_solution}
\end{equation}

\subsection{RGBD and SfM Depth Fusion\label{subsec:Fusing-RGBD-Depths}}

For fusing measurements we consider the structured-light measurement
of the RGBD sensor to generate a distribution for the unknown true
depth of the scene surfaces at each $(x,y)$ pixel in the depth image.
These measurements are considered to be independent and identically
distributed to the measurements of the true unknown depth of the scene
surfaces from the registered SfM estimated depths. With these assumptions,
solving the depth fusion problem is equivalent to estimating the posterior
distribution of the true scene depth at each $(x,y)$ position given
the distributions for the RGBD and SfM depth values.

Fortunately, previous sections show that Gaussian models are appropriate
distributions for both the RGBD and SfM depth values and the parameters
of these models are either known (see \S~\ref{subsec:RGBD_Measurement-Noise})
or estimated continuously (see \S~\ref{subsec:Time-and-Spatial}).
When both distributions are Gaussian, the posterior distribution can
be found analytically and is a well-known result used in pattern recognition
and other prediction frameworks, e.g., the Kalman filter as discussed
in \cite{Maybeck79stochasticsmodels}. Specifically, let the Gaussian
noise for RGBD depth at position $(x,y)$ be represented as $\mathcal{N}(\mathbf{d}_{rgbd},\sigma_{rgbd}^{2})$
and the Gaussian noise for the co-registered SfM depth image at position
$(x,y)$ be $\mathcal{N}(\mu_{SfM},\sigma_{SfM}^{2})$. The posterior
distribution on the unknown true depth at position $(x,y)$ is also
Gaussian and let $\mathcal{N}(\mu_{fused},\sigma_{fused}^{2})$ denote
the mean and variance parameters of this distribution. Equations (\ref{eq:fused_mean})
and (\ref{eq:fused_variance}) provide optimal estimates of the mean
and variance of the fused depth at position $(x,y)$. The best estimate
of the fused depth is given by the highest probability value in the
posterior distribution which is the mean fused image, $\mu_{fused}(x,y)$.

\begin{equation}
\mu_{fused}=\frac{\mathbf{d}_{rgbd}\sigma_{SfM}^{2}+\mu_{SfM}\sigma_{rgbd}^{2}}{\sigma_{SfM}^{2}+\sigma_{rgbd}^{2}}\label{eq:fused_mean}
\end{equation}

\begin{equation}
\sigma_{fused}^{2}=\frac{\sigma_{rgbd}^{2}\sigma_{SfM}^{2}}{\sigma_{rgbd}^{2}+\sigma_{SfM}^{2}}\label{eq:fused_variance}
\end{equation}

\section{Results}

\begin{figure*}
\noindent \begin{centering}
\subfloat[]{\noindent \begin{centering}
\begin{minipage}[t]{1.3in}%
\noindent \begin{center}
\includegraphics[height=1in]{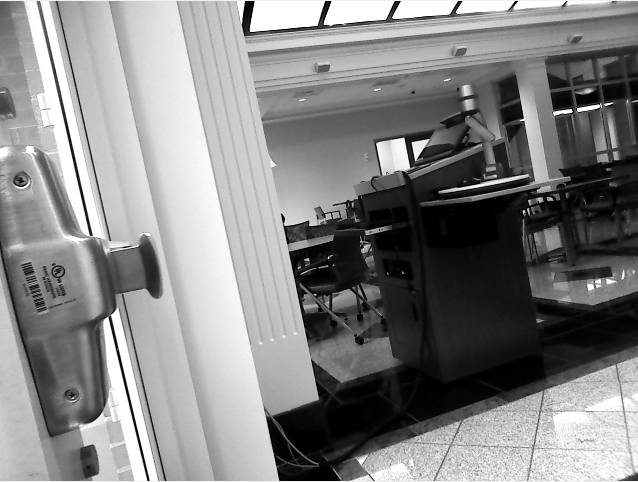}
\par\end{center}
\noindent \begin{center}
\includegraphics[height=1in]{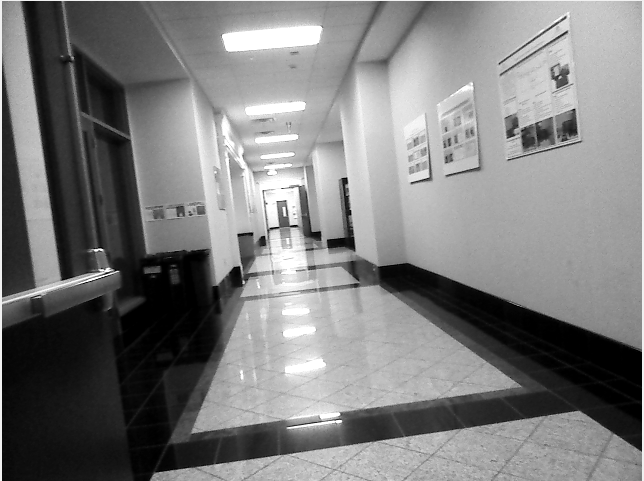}
\par\end{center}
\noindent \begin{center}
\includegraphics[height=1in]{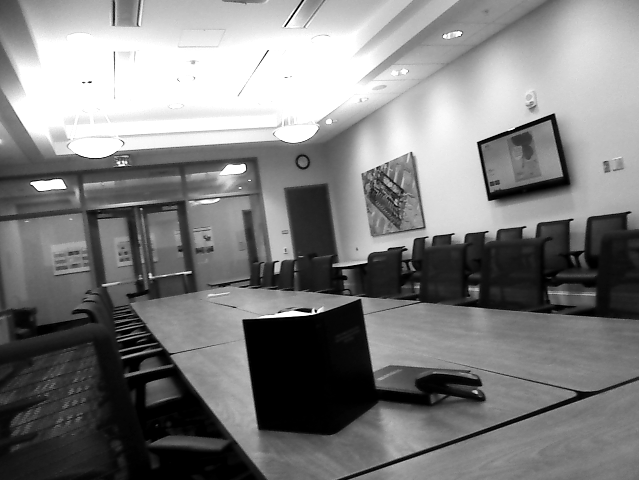}
\par\end{center}%
\end{minipage}
\par\end{centering}
}\subfloat[]{\noindent \begin{centering}
\begin{minipage}[t]{1.3in}%
\noindent \begin{center}
\includegraphics[height=1in]{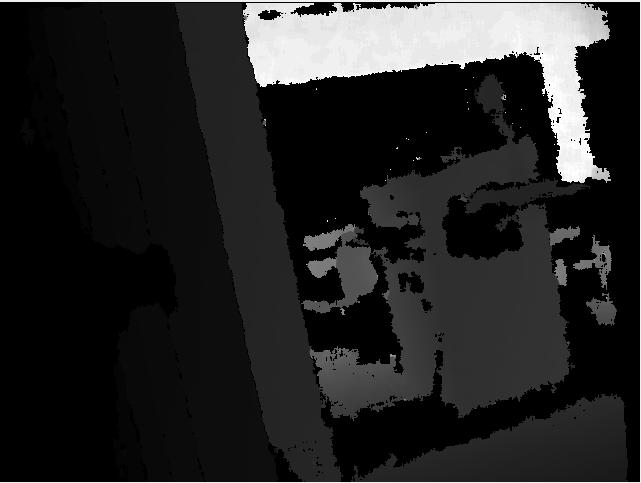}
\par\end{center}
\noindent \begin{center}
\includegraphics[height=1in]{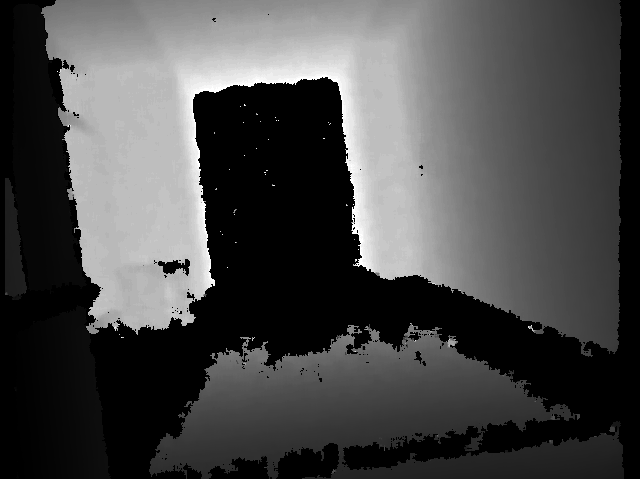}
\par\end{center}
\noindent \begin{center}
\includegraphics[height=1in]{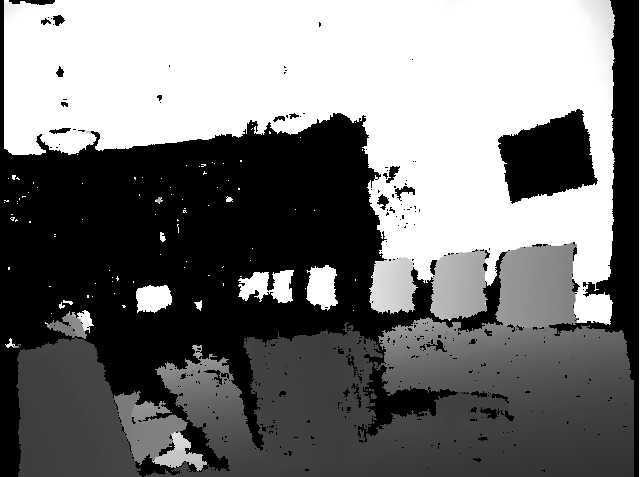}
\par\end{center}%
\end{minipage}
\par\end{centering}
}\subfloat[]{\noindent \begin{centering}
\begin{minipage}[t]{1.3in}%
\noindent \begin{center}
\includegraphics[height=1in]{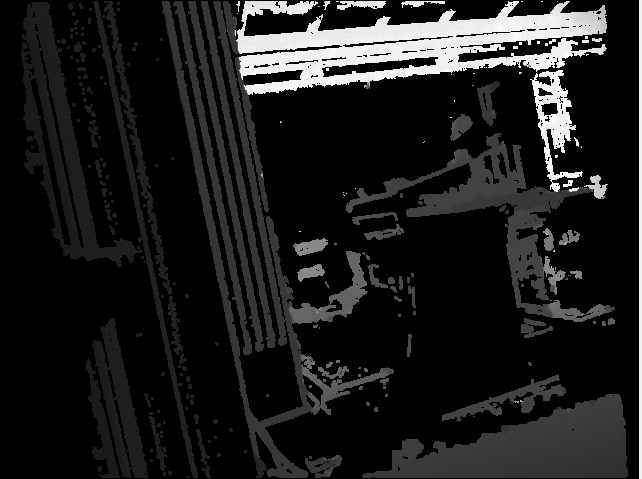}
\par\end{center}
\noindent \begin{center}
\includegraphics[height=1in]{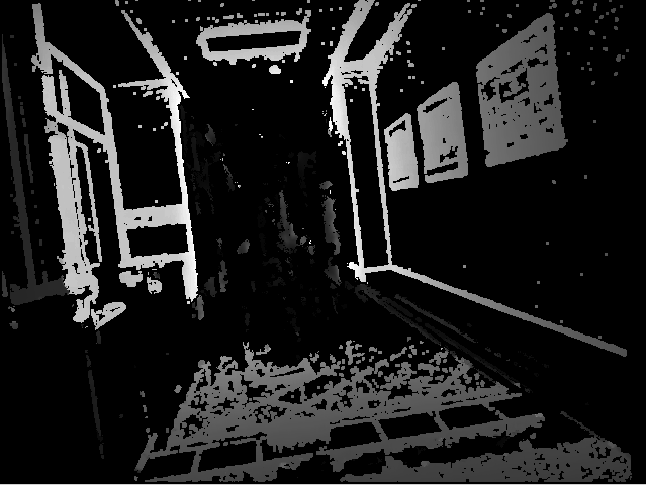}
\par\end{center}
\noindent \begin{center}
\includegraphics[height=1in]{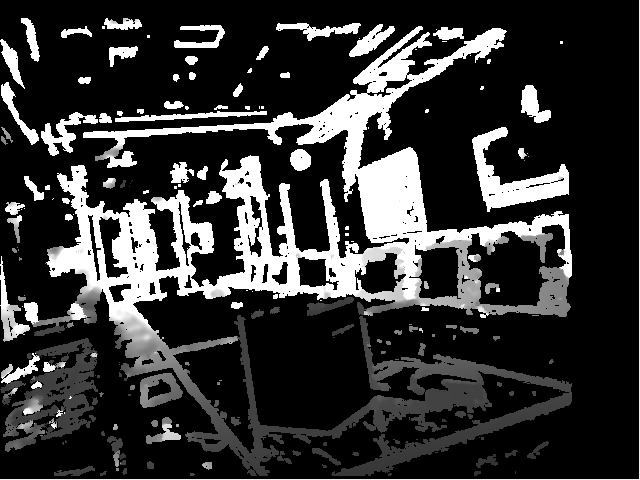}
\par\end{center}%
\end{minipage}
\par\end{centering}
}\subfloat[]{\noindent \begin{centering}
\begin{minipage}[t]{1.3in}%
\noindent \begin{center}
\includegraphics[height=1in]{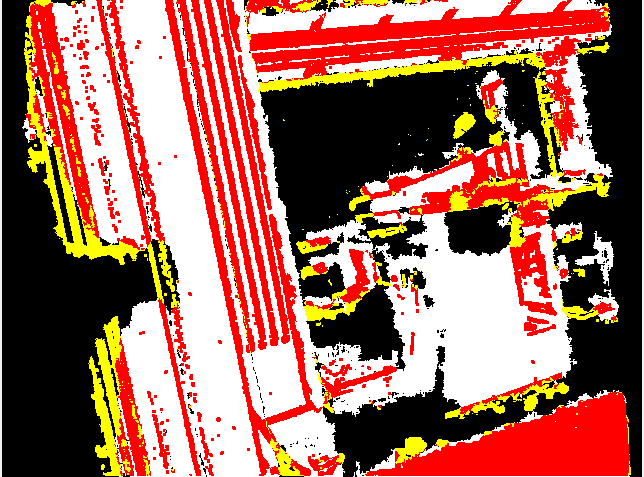}
\par\end{center}
\noindent \begin{center}
\includegraphics[height=1in]{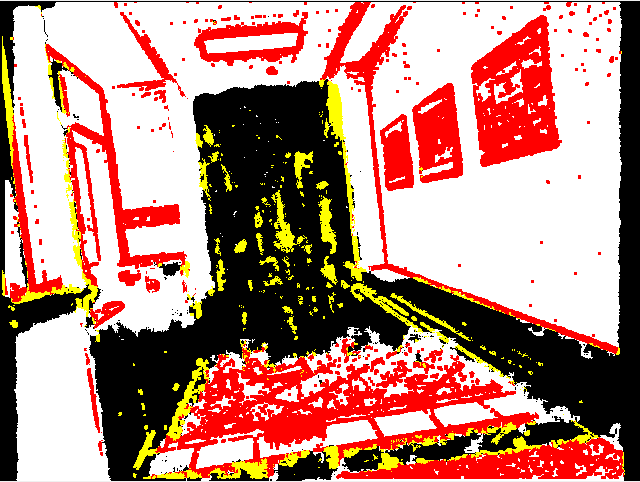}
\par\end{center}
\noindent \begin{center}
\includegraphics[height=1in]{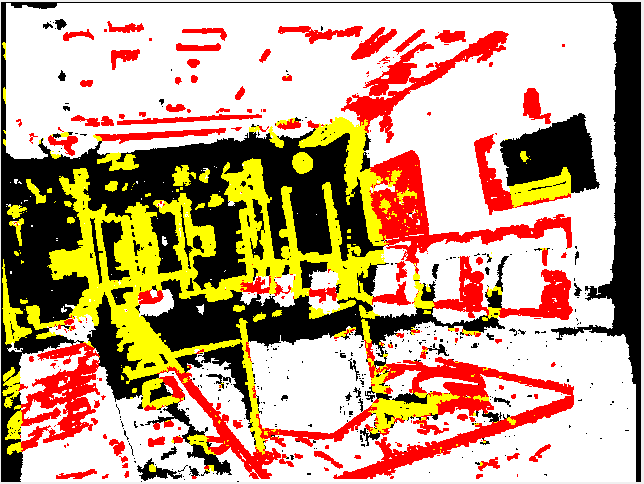}
\par\end{center}%
\end{minipage}
\par\end{centering}
}\subfloat[]{\noindent \begin{centering}
\begin{minipage}[t]{1.1in}%
\noindent \begin{center}
\includegraphics[height=1in]{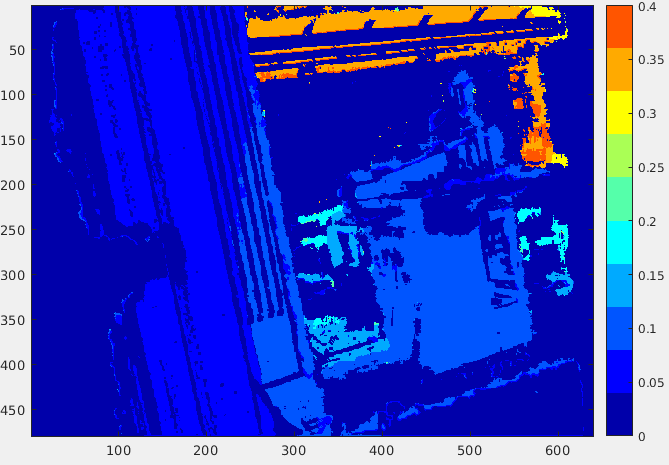}
\par\end{center}
\noindent \begin{center}
\includegraphics[height=1in]{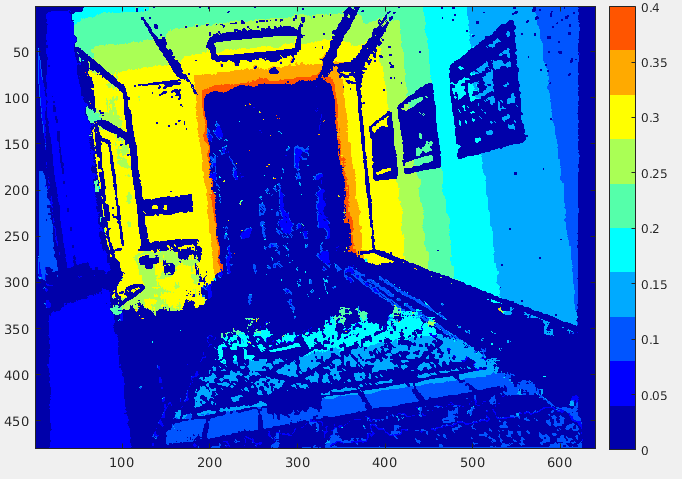}
\par\end{center}
\noindent \begin{center}
\includegraphics[height=1in]{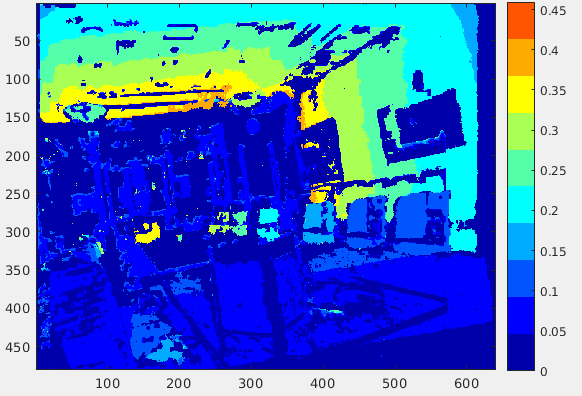}
\par\end{center}%
\end{minipage}
\par\end{centering}
}
\par\end{centering}
\noindent \centering{}\caption{Results for three experiments are shown. Images shown are organized
into separate columns. Column (a) shows a grayscale image of the scene
(b) shows the sensed RGBD depth image (c) shows the SfM-estimated
depth image (d) shows the fused image and (e) shows the standard deviation
for fused depths (in $m.$). The fused image has been color-coded
as follows: (white) denotes depth locations sensed only by the RGBD
sensor, (yellow) denotes depth locations only sensed via SfM, (red)
denotes fused (RGBD+SfM) depth locations and (black) denotes depth
locations without RGBD or SfM measurements.}
\end{figure*}
We have applied the depth fusion algorithm to sets of RGBD images
of a variety of indoor scenes. RGBD image sequences were recorded
using an Orbbec Astra RGBD sensor. The LSD-SLAM algorithm was applied
to the RGB image sequence using the factory provided intrinsic camera
calibration parameters. Each sequence included approximately 60 seconds
of RGBD image data at the rate of 30 fps. Experimentally recorded
RGB images were processed offline by the LSD-SLAM algorithm to generate
SfM depth images. Experiments initialize the LSD-SLAM algorithm with
the first recorded depth image from the RGBD sensor to facilitate
the initial scale approximation. The output from the LSD-SLAM algorithm
consisting of the relative pose for every tracked frame and the depth
map for each keyframe was then recorded. The fusion algorithm was
then applied using the recorded RGBD image stream and LSD-SLAM output
files to generate the results shown in this section.
\begin{table}
\noindent \begin{centering}
\begin{tabular}{|c|c|c|c|}
\hline 
\multirow{2}{*}{{\footnotesize{}Experiment}} & {\footnotesize{}RGBD-only depths} & {\footnotesize{}SfM-only depths} & {\footnotesize{}Fused depths}\tabularnewline
 & {\footnotesize{}(\%)} & {\footnotesize{}(\%)} & {\footnotesize{}(\%) }\tabularnewline
\hline 
\hline 
1 & 67.7 & 6.7 & 25.5\tabularnewline
\hline 
2 & 55.5 & 10.0 & 34.3\tabularnewline
\hline 
3 & 68.1 & 15.1 & 16.6\tabularnewline
\hline 
\hline 
{\footnotesize{}Average} & 63.3 & 10.3 & 26.3\tabularnewline
\hline 
\end{tabular}
\par\end{centering}
\caption{Ratios of depth measurements to total depths\label{tab:Ratios-of-depth}}
\end{table}

Experiment 1 depicts an indoor office scene at the university. This
scene includes specular and dark surface structures at close range
that are not measured by the RGBD sensor. Yet, the SfM algorithm estimates
depths at a number of locations (on the podium) where there are significant
intensity changes. These additional depths are evident in the fused
results, which includes SfM-only depth measurements in regions in
the vicinity of image edges. The experiment demonstrates that depth
fusion can improve depth images by obtaining depths from surfaces
not measurable by the RGBD sensor.

Experiment 2 depicts a hallway in the UNCC EPIC building that includes
both specular surfaces and high intensity illumination from overhead
lights. The experiment is a second example showing that depth fusion
bolsters over all depth measurement performance by providing the depths
when RGBD sensors fail. The SfM takes advantage of the patterns on
the surface and estimates the depths irrespective of the nature of
its reflectance properties and color.

Experiment 3 depicts the UNCC faculty conference hall. Here a number
of scene structures lie beyond the measurement range of the RGBD sensor.
Yet, the SfM algorithm is able to estimate the depth of these scene
structures (albeit at high variance) providing depths that would otherwise
not be possible.

Table \ref{tab:Ratios-of-depth} quantifies the amount of additional
depth information provided via RGBD-SfM depth image fusion. On average,
approximately 63\% of the fused depths originate from the RGBD sensor
alone, approximately 10\% original from the SfM algorithm alone, and
approximately 26\% of the fused depths results from RGBD-SfM fusion.
The addition of 10\% of novel depth data is a significant contribution.
Further, fused depths account for roughly 26\% of depth image data
and the measurement error for all of these measurements will be reduced
by the fusion. Variance reduction will be greatest for low-variance
SfM depth estimates which are typically in textured scene locations
close to the camera. Yet, we note that by inspection of equation (\ref{eq:fused_variance}),
it is theoretically impossible for the variance of any fusion result
to increase.

\section{Conclusions}

The SfM depth estimation complements the RGBD sensor measurements
and can provide depths when RGBD sensors fail. The proposed depth
fusion algorithm provides an effective way to augment the RGBD depth
stream and results in improved depth images. The experiments conducted
show improvements in the resulting depth images for a number of scenarios
where structured light depth estimation fails. This includes successfully
capturing depth for out-of-range RGBD depth locations and successfully
capturing depth measurements from specular and dark objects. 

In future work, the proposed depth fusion method can be generalized
to address depth image locations that include measurements having
non-Gaussian noise distributions. Also, experimental results show
that the scale and trajectory estimation of SfM depths are not always
accurate. Inclusion of RGBD-SfM fused depth data into the SfM algorithm
can be exploited to improve the camera pose estimation and increase
reconstruction accuracy.

\bibliographystyle{ieeetr}
\bibliography{references_db,references_db_willis}

\end{document}